\documentclass[12pt]{article}
\usepackage[utf8]{inputenc}
\usepackage[T1]{fontenc}

\usepackage[compress]{cite}
\usepackage{IEEEtrantools,stackengine}
\stackMath

\usepackage{authblk}
\usepackage{fullpage}
\usepackage{amssymb,amsmath}
\usepackage{adjustbox}
\usepackage{graphicx}
\usepackage{longtable}

\usepackage{siunitx}
\usepackage{csquotes}

\usepackage{xspace}

\usepackage{booktabs}
\usepackage{longtable}
\usepackage{multirow}

\usepackage[dvipsnames, table]{xcolor}

\usepackage{tikz} 
\usepackage{pgf-pie}
  \usetikzlibrary{patterns}

\newcommand{\PIE}[1]{\begin{tikzpicture}
\pie[ 
  color = {black!0, darkgreen!80},radius = .25, rotate = 90,
  /tikz/nodes={text opacity=0,overlay}
] {\the\numexpr100-#1\relax, #1}
\end{tikzpicture}}

\usetikzlibrary{arrows.meta}

\usepackage[left]{lineno}

\definecolor{darkgreen}{rgb}{0.0, 0.5, 0.0}

\definecolor{lightyellow}{HTML}{FFE699}
\definecolor{red_revision}{HTML}{FF0000}

\usepackage{setspace}
\usepackage[unicode=true]{hyperref}
\hypersetup{breaklinks=true,
            pdfborder={0 0 0},
            allcolors=black}
\usepackage[%
  sort&compress
]{cleveref}
  \Crefname{appendix}{Supplement}{Supplements}
  \Crefname{figure}{Fig.}{Fig.}

\usepackage{times}
\usepackage{enumerate}
\usepackage{enumitem}
\usepackage{float}

\usepackage{subfig}
\usepackage{graphicx}

\usepackage{rotating}
\usepackage{tabularx}
\usepackage{dcolumn}
\usepackage{pdflscape}
\usepackage{rotating}

\usepackage{array}

\usepackage{titlesec}

\titleformat{\subsubsection}
  {\normalfont\itshape}{\thesubsubsection}{1em}{}

\titleformat{\subsection}
  {\normalfont\bfseries}{\thesubsection}{1em}{}

\makeatletter
\renewcommand{\fps@figure}{H}         
\renewcommand{\fps@table}{H}         
\makeatother

\allowdisplaybreaks

\usepackage{eurosym}
\usepackage{comment}

\usepackage{ntheorem}
\theoremseparator{:}

\usepackage{makecell}

\begin{document}


\title{\vspace{-1cm}\centering\LARGE\singlespacing Causal machine learning for predicting treatment outcomes}

\renewcommand\Authfont{\fontsize{9}{10.8}\selectfont}
\renewcommand\Affilfont{\fontsize{9}{10.8}\selectfont}

\author[,1,2]{Stefan Feuerriegel\thanks{Corresponding author: feuerriegel@lmu.de}}
\author[1,2]{Dennis Frauen}
\author[1,2]{Valentyn Melnychuk}
\author[1,2]{Jonas Schweisthal}
\author[1,2]{Konstantin Hess}
\author[3]{Alicia Curth}
\author[4,5]{Stefan Bauer}
\author[2,4,5]{Niki Kilbertus}
\author[6]{Isaac S. Kohane}
\author[7,8]{Mihaela van der Schaar}

\affil[1]{LMU Munich, Munich, Germany}
\affil[2]{Munich Center for Machine Learning, Munich, Germany}
\affil[3]{Department of Applied Mathematics \& Theoretical Physics, University of Cambridge, Cambridge, United Kingdom}
\affil[4]{School of Computation, Information and Technology, TU Munich, Munich Germany}
\affil[5]{Helmholtz Munich, Munich, Germany}
\affil[6]{Department of Biomedical Informatics, Harvard Medical School, Boston, USA}
\affil[7]{Cambridge Centre for AI in Medicine, University of Cambridge, Cambridge, United Kingdom}
\affil[8]{The Alan Turing Institute, London, United Kingdom}

\date{}

\maketitle

\vspace{-1cm}

\begin{abstract}\normalfont
\noindent
{Causal machine learning (ML) offers flexible, data-driven methods for predicting treatment outcomes. Here, we present how methods from causal ML can be used to understand the effectiveness of treatments, thereby supporting the assessment and safety of drugs.} A key benefit of causal ML is that allows for estimating individualized treatment effects, as well as personalized predictions of potential patient outcomes under different treatments. This offers granular insights into when treatments are effective, so that decision-making in patient care can be personalized to individual patient profiles. {We further discuss how causal ML can be used in combination with both clinical trial data as well as real-world data such as clinical registries and electronic health records.} We finally provide recommendations for the reliable use of causal ML in medicine. 
\end{abstract}

\noindent {\scriptsize First published in Nature Medicine, 30, 958–968 (2024) by Springer Nature. Link: \url{https://doi.org/10.1038/s41591-024-02902-1}}

\flushbottom
\maketitle
\thispagestyle{empty}

\sloppy
\raggedbottom

\newpage
\section*{Main}
\label{sec:introduction}

Assessing the effectiveness of treatments is crucial to ensure patient safety and personalize patient care. {Recent innovations in machine learning (ML) offer new, data-driven methods to estimate treatment effects from data. This branch in ML is commonly referred to as causal ML as it aims to predict a causal quantity, namely, the patient outcomes due to treatment \cite{Kaddour.2022}. Causal ML can be used in order to estimate treatment effects from both experimental data obtained through randomized controlled trials (RCTs) and observational data obtained from clinical registries, electronic health records, and other real-world data (RWD) sources to generate clinical evidence.} A key strength of causal ML is that it allows to estimate individualized treatment effects, as well as to make personalized predictions of potential patient outcomes under different treatments. This offers a granular understanding of when treatments are effective {or harmful}, so that decision-making in patient care can be personalized to individual patient profiles.

\clearpage

\noindent\fbox{%
\parbox{\textwidth}{%
\color{black}
\textbf{Box 1. Glossary of common terms in causal ML}

\singlespacing
\begin{flushleft}
\begin{itemize}
\item Causal graph: A graphical representation of the causal relationships between variables, typically using directed acyclic graphs to depict causal paths.
\item Causal ML: A branch of machine learning that aims at the estimation of causal quantities (e.g., average treatment effect, conditional average treatment effect) or at predicting potential outcomes. Here, ``causal'' implies that the target is a causal quantity when certain assumptions about the data-generating mechanism are satisfied. For alternative definitions and use cases of causal ML, see \cite{Kaddour.2022}.
\item Confounder: A variable that influences both the treatment assignment and the outcome.
\item Consistency: The potential outcome is equal to the observed patient outcome under the selected treatment, which implies that the outcomes are clearly defined.
\item Counterfactual outcome: The unobservable patient outcome that would have occurred, had a patient received a different treatment.
\item Factual outcome: The observed patient outcome that occurred for the observed treatment.
\item Identifiability: A statistical concept referring to the ability of
whether causal quantities such as treatment effects
can be uniquely inferred from the observed data.
\item Positivity: Each patient has a bigger-than-zero probability of receiving/not receiving a treatment. This is also called overlap assumption.
\item Potential outcome: The hypothetical patient outcome that would be observed if a certain treatment was administered.
\item Propensity score: The propensity score is the probability of receiving the treatment given the observed specific patient characteristics.
\item Stable unit treatment value (SUTVA): The outcome for any patient does not depend on the treatment assignment of other patients, and there is no variation in the effect of the treatment across different settings or populations.
\item Unconfoundedness: Given observed covariates, the assignment to treatment is independent of the potential outcomes. This is the case e.g., when there are no unobserved confounders, that is, variables influencing both the treatment and the outcome. The assumption is also called ignorability.
\end{itemize}
\end{flushleft}
}}

\begin{figure}
\qquad\textbf{\textsf{a}}
\begin{center}
\vspace{-1cm}
\includegraphics[width=.6\linewidth]{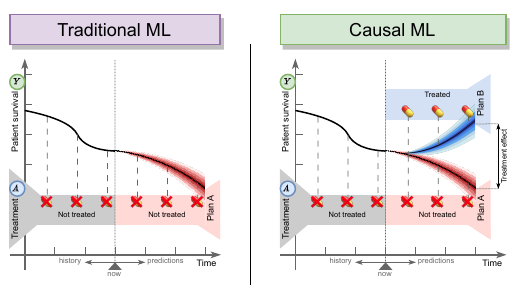}
\end{center}

\qquad\textbf{\textsf{b}}
\begin{center}
\vspace{-1cm}
\includegraphics[width=.6\linewidth]{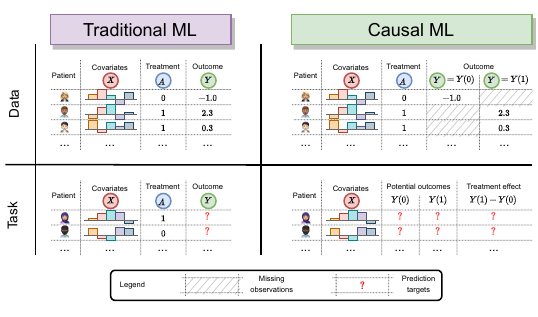}
\end{center}

\qquad\textbf{\textsf{c}}
\begin{center}
\vspace{-1cm}
\includegraphics[width=.4\linewidth]{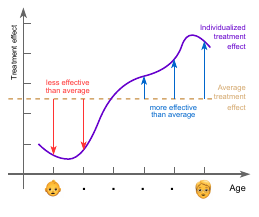}
\end{center}

\vspace{-.1cm}

\caption{\textbf{Causal ML for predicting treatment outcomes.} \textbf{a},~Different from traditional ML, causal ML aims at (i)~estimating the treatment effect or (ii)~predicting the patient outcomes themselves due to treatments. \textbf{b},~Causal ML is challenging due to the fundamental problem of causal inference in that not all potential outcomes can be observed and are thus missing in the data. \textbf{c},~Treatment effect heterogeneity. }
\label{fig:differences_causal_ML}
\end{figure}

\section*{Causal ML in medicine}

\subsection*{Comparison to traditional ML}

{Causal ML for estimating treatment effects is different from traditional predictive ML (see Box~1 for a glossary of terms).} Intuitively, traditional ML aims at predicting outcomes \cite{Esteva.2019}, while causal ML quantifies changes in outcomes due to treatment, so that treatment effects can be estimated (see \Cref{fig:differences_causal_ML}a). For example, a typical use case for traditional ML is risk scoring,  such as predicting the probability of a diabetes onset to understand which patients are at high risk but without saying what the best treatment plan is \cite{Kopitar.2020,Alaa.2019cardiovascular,Cahn.2020,Zueger.2022,Krittanawong.2020}. In contrast, causal ML aims to answer ``what if'' questions. For example, causal ML could answer the question of how the risk of a diabetes onset will \emph{change} if the patient receives an antidiabetic drug \cite{Xie.2023,Deng.2023,Kalia.2023}, so that decisions can be made whether to administer an antidiabetic drug. Causal ML can also be used to predict the potential patient outcomes in response to different treatments. For example, in oncology, causal ML could make individualized predictions of survival under different treatment plans, which can then help medical practitioners in choosing a treatment plan that promises the largest chance of survival \cite{Petito.2020}.

Estimating treatment effects from data requires custom methods. The reason is that treatment effects for individual patients are not observable. This is due to the so-called fundamental problem of causal inference \cite{Holland.1986, Pearl.2009}: one can only observe the (factual) patient outcome under the given treatment, but one never observes the (counterfactual) patient outcome under a different, hypothetical treatment (see \Cref{fig:differences_causal_ML}b). Therefore, the estimation of treatment effects or other causal quantities, which are based on such unobserved outcomes, poses additional challenges not present in traditional, predictive ML.

{To then obtain a causal quantity that can be estimated, certain assumptions on the causal structure of the problem must be made. In particular, one often needs to assume that there is no unmeasured confounding, that is, there are no unobserved factors that drive both treatments and patient outcomes.} If {unmeasured} confounding is present, the estimated treatment effects may suffer from confounding bias and, as a result, can even have a wrong sign \cite{Hemkens.2018}. Additionally, to estimate treatment effects, one needs to account for the dependence structure between treatment, outcomes, and patient characteristics by modeling the underlying causal relationships. One reason is that intervening on the treatment variable could also affect other patient characteristics. As an example, consider a patient with a certain body mass index for whom the diabetes risk should be predicted and where the doctor recommends stopping smoking. Literature from traditional ML would suggest using both the body mass index and smoking behavior to predict the diabetes risk under smoking vs. no smoking; however, this approach would ignore that stopping smoking would also change a patient's body mass index. {Instead, ML needs to be embedded in a causal framework.}

\subsection*{Benefits}

{Methods for estimating treatment effects have a long tradition in the statistics literature (e.g., \cite{Rubin.1974, Rubin.2005, Robins.1994, Robins.1999}). Causal ML builds upon the same problem setup but makes changes to the estimation strategy. Hence, the core improvement from using causal ML is generally not the types of questions that can be asked, but how these questions can be answered. As such, causal ML can have benefits over alternative methods from the statistics literature (see Box~2).} First, methods from classical statistics often assume knowledge about the parametric form of the association between patient characteristics and outcomes, such as linear dependencies. However, such knowledge is often not available or unrealistic, especially for high-dimensional datasets such as electronic health records, and this can easily lead to models that are misspecified. {In contrast, causal ML typically allows for less rigid models, which helps in capturing complex disease dynamics as well as human pathophysiology and pharmacology. Still, there is a trade-off as causal ML typically requires larger sample sizes. For example, parsimonious models such as linear regressions are often favorable for settings with small sample sizes, while more flexible, non-linear models are only effective for large sample sizes.}

\noindent\fbox{%
\parbox{\textwidth}{%
\color{black}
\textbf{Box 2. Comparison of causal ML vs. traditional statistics}

\begin{flushleft}
Due to the importance of treatment effect estimation across many application areas, methods for treatment effect estimation have been developed in different disciplines, including statistics, biostatistics, econometrics, and machine learning (e.g., \cite{vanderLaan.2006, vanderLaan.2011tmle, Foster.2023,Kunzel.2019, Curth.2021nonparametric, Kennedy.2023DR, Nie.2021quasioracle, Chernozhukov.2018}). However, there is no `dichotomy' as many concepts are shared across disciplines. For example, many state-of-the-art methods for estimating treatment effects are {model-agnostic} in that they can be used in combination with both arbitrary models from classical statistics and also more modern machine learning models \cite{vanderLaan.2006, vanderLaan.2011tmle, Foster.2023}. 

\vspace{0.3cm}

Eventually, the choice of whether to rely on a more classical statistical model or a more modern machine learning method presents a trade-off that depends on the underlying settings. For example, simple models (e.g., linear regression or other parametric models) are often preferred for small sample sizes. For large sample sizes, more complex, non-linear models can be used to capture heterogeneity in the treatment effect. Notwithstanding, the ability to handle non-linear relationships and treatment effect heterogeneity is not unique to causal ML but can, in principle, also rely on classical statistical models that allow incorporating prespecificed non-linearities. Therefore, causal ML may have advantages when the underlying data-generating process is complex and when prior knowledge is limited. 

\end{flushleft}

}}

\vspace{0.5cm}

In medicine, causal ML offers several opportunities for estimating treatment effects from data, which eventually help in greater personalization of care. First, at the patient level, causal ML can handle high-dimensional and unstructured data with patient covariates and thus estimate treatment effects from multi-modal datasets containing images, text, or time series, as well as genetic data. For example, one could estimate treatment effects from computed tomography (CT) scans or entire electronic health records. Second, at the outcome level, causal ML can help make personalized estimates of treatment effects for subpopulations or even predict outcomes for individual patients \cite{Yoon.2018}. For example, individual differences in drug metabolism can lead to serious side effects for drugs in some patients but can be lifesaving in others \cite{Evans.1999}, so causal ML could learn such differences and thus help in designing personalized treatment strategies. Third, at the treatment level, causal ML can be effective for estimating heterogeneity in treatment effects across patients in a data-driven manner to better understand where treatments are effective (see \Cref{fig:differences_causal_ML}c).

\section*{Workflow}

\begin{figure}
\begin{center}
\includegraphics[width=.5\linewidth]{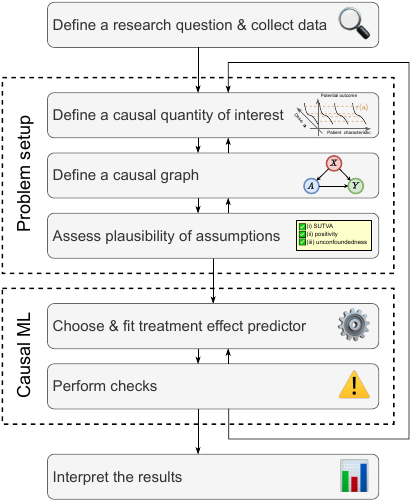}
\end{center}
\caption{\textbf{Workflow for causal ML in medicine.}}
\label{fig:workflow}
\end{figure}

The workflow in \Cref{fig:workflow} outlines the different steps necessary to predict treatment outcomes with causal ML. The workflow {\cite{Dang.2023,Petersen.2014}} should help researchers in clearly defining the research question and then guide the choice of the causal quantity of interest, the causal model, the causal ML method, and the robustness checks to validate the reliability of the estimates.

\subsection*{Problem setup}

\begin{figure}
\qquad\textbf{\textsf{a}}
\begin{center}
\vspace{-1cm}
\includegraphics[width=.5\linewidth]{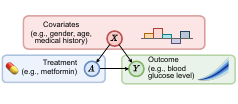}
\vspace{-0.5cm}
\end{center}

\qquad\textbf{\textsf{b}}
\begin{center}
\includegraphics[width=\linewidth]{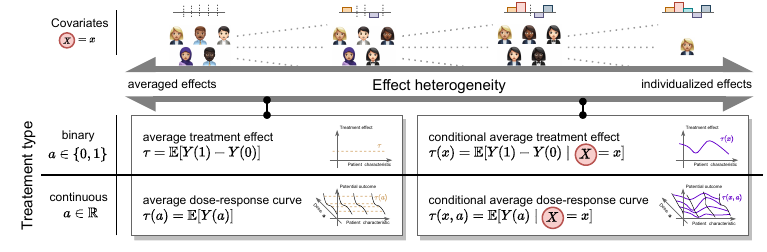}
\end{center}

\textbf{\textsf{c}}
\begin{center}
\includegraphics[width=0.9\linewidth]{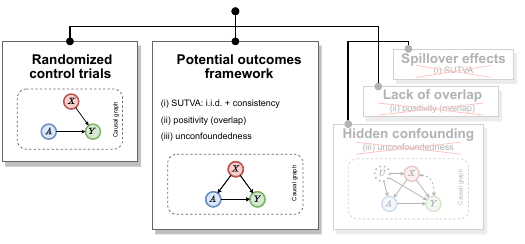}
\end{center}
\caption{\textbf{Formalizing tasks for causal ML.} \textbf{a},~A causal graph must be assumed such as the example here. The arrows ($\rightarrow$) indicate the causal relationships between different variables. Note that the causal graph allows for possible unobserved variables (not confounders) that are correlated with treatment and confounders, or correlated with confounders and outcome. \textbf{b},~The research question defines what causal quantity is of interest, that is, the so-called estimand. The estimand can vary by the effect heterogeneity (average vs. individualized) and treatment type (binary vs. continuous). Here, $Y(a)$ is the potential outcome for treatment $a$. \textbf{c},~The research question determines the problem setup, which comes with different assumptions that must be made to ensure identifiability and thus reliable inferences.}
\label{fig:setup}
\end{figure}

To estimate the effectiveness of treatments, information about the following variables is necessary \cite{Pearl.2009}: the treatment of interest; the observed patient outcome; and patient characteristics such as age, gender, and the medical history. The patient characteristics are commonly also called covariates. For example, in cancer care, one could use electronic patient records with information about the type of chemotherapy (the treatment), the size of a cancer tumor (the outcome), and the previous medical history (the covariates). In the standard setting \cite{Pearl.2009}, the variables can influence each other as shown by the example causal graph in \Cref{fig:setup}a. To make causal quantities identifiable, we later need to assume knowledge about the causal graph. 

{Information about the above variables can come from either observational or experimental data.} In observational data, the treatment assignment is unknown and not fully randomized. This is the case in RWD such as clinical registries and electronic health records. Here, the treatment assignment followed some, typically unknown procedure depending on the patient characteristics. For example, patients with a more severe illness are likely to get a more aggressive form of treatment, implying that the patient characteristics differ across treatment groups. This is unlike RCTs, where treatments are randomized, which is also referred to as experimental data. As a result, the patient characteristics are similar across treatments. This is captured by the propensity score, which is the probability of receiving a treatment given the patient covariates \cite{Rubin.1974}. In RCTs, the propensity score is known (e.g., the propensity score is 50\% in completely randomized trials with two treatment arms of equal size). In contrast, the treatment assignment in RWD is unknown but can be estimated to account for differences in the patient populations of those who have received treatment and those who have not. 

\newpage
\thispagestyle{empty}

\subsection*{Types of treatment effects}

{Outcomes of treatments are commonly formalized based on the potential outcomes framework \cite{Rubin.1974}. The framework conceptualizes so-called potential outcomes, which are the patient outcomes that would hypothetically be observed if a certain treatment was administered. Then, depending on the practical applications, different causal quantities can be of interest.} These include treatment effects, which quantify the expected difference of two potential outcomes under different treatments. Common choices of treatment effects can be loosely grouped along two dimensions (see \Cref{fig:setup}b): (i)~the degree of effect heterogeneity and (ii)~the treatment type. By choosing a specific treatment effect of interest, one defines the so-called estimand, that is, the causal quantity that should be estimated by the causal ML method. 

(i)~\emph{Degree of effect heterogeneity}: Traditionally, average treatment effects (ATEs) are widely used in clinical trials \cite{Rubin.1974}. ATEs measure effects at the level of the study population. By comparing the average patient outcome for those receiving the treatment versus those who do not (control group), ATEs help in understanding how effective a treatment is, on average, across a specific patient cohort \cite{vanderLaan.2006}. This is important, for example, when analyzing the comparative effectiveness of a new drug compared to the standard of care, or when assessing the overall effectiveness or safety of a new drug. However, ATEs can not offer granular insights into whether patients with specific covariates may particularly benefit from a treatment. Nevertheless, such heterogeneity in treatment effects can be of high interest in practice (see \Cref{fig:differences_causal_ML}c). For this, one typically estimates the conditional average treatment effect (CATE), which is the effect of a treatment for a particular subgroup of patients defined by the covariates. Understanding the heterogeneity in treatment effects informs about subgroups where treatments are not effective or even harmful, which is relevant for individualizing treatment recommendations to specific patients.

(ii)~\emph{Treatment type}: Binary treatments refer to a type of treatment variable that is dichotomous and thus has only two categories, for example, when answering questions of whether to treat or not to treat. In contrast, {discrete as well as continuous treatments} refer to a type of treatment variable that can take on a range of values rather than being limited to two (or a few) categories. Continuous treatment variables are commonly present in situations where the intensity, dosage, or level of exposure to treatment can be flexibly chosen \cite{Hirano.2004}. For example, in radiation therapy, the dose of radiation is often chosen from a fairly wide spectrum that depends on the cancer type and other patient characteristics \cite{Specht.2014}. For continuous treatments, the treatment effectiveness is often also summarized by dose-response curves.

{Besides the above, some applications in medicine are also interested in the potential outcomes themselves, rather than in treatment effects. While understanding the treatment effectiveness is often important for the assessment and safety of drugs, the potential outcomes can support decision-making in routine care by helping clinical professionals reason about what outcome to expect under different treatment options. This may be seen as a `risk under intervention' estimand and requires a careful modeling strategy \cite{vanGeloven.2020}. For example, while the treatment effect may say that a drug can improve the 5-year mortality by 5 percentage points, the predicted outcomes inform that the mortality is 20\% with treatment and 15\% without, so that not only the relative gain but the future health state can be considered. However, in practice, the estimation of ATE and CATE is often an easier task than estimating potential outcomes \cite{Kennedy.2023DR} and, hence, is preferred when it is sufficient for decision-making.}

\subsection*{Assumptions for identifiability}

The estimation of treatment effects involves counterfactual outcomes, which are not observable. Therefore, formal assumptions must be made about the data-generating process to ensure the identifiability of treatment effects from data \cite{Pearl.2009}. Intuitively, identifiability is a theoretical concept that refers to whether causal quantities such as treatment effects can be uniquely inferred from data. Ensuring identifiability is necessary since, otherwise, it is impossible in practice to estimate a treatment effect without bias even with infinite data \cite{Pearl.2009}.

RCTs ensure the identifiability of treatment effects through fully randomized treatment assignment. However, treatment assignment in RWD is unknown and not fully randomized, namely, it depends on other covariates, so that formal assumptions must be made \cite{Rubin.1974}. The exact set of assumptions depends on which type of treatment effect is chosen. For the treatment effects discussed above, the following three assumptions (see \Cref{fig:setup}c) are standard \cite{Rubin.1974,Imbens.2015}: (i)~Stable unit treatment value assumption (SUTVA) implies that the RWD are independent and identically distributed and that consistency holds (i.e., the potential outcome coincides with the observed outcome for a given treatment). This assumption implies that there is no interference between patients in which treating one patient influences the outcomes for another patient in the study population (e.g., due to spillover or peer effects). (ii)~Positivity (also called overlap) requires a non-zero probability of receiving a treatment. Positivity implies that, for each possible combination of patient characteristics, we can observe both treated and untreated patients. (iii)~Unconfoundedness (also called ignorability) states that, given observed covariates, the treatment assignment is independent of the potential outcomes. In particular, this is satisfied if the patient covariates include all possible confounders, i.e., variables that influence both the treatment and the outcome. For example, unconfoundedness may be violated if patients with certain sociodemographics such as race or high income tend to have better access to treatments and better adherence \cite{Chen.2016} and where the reason is not captured in the data. In principle, unconfoundedness can be addressed by capturing all relevant factors behind driving treatment assignment in RWD \cite{Cinelli.2022}, yet it is generally challenging to validate this in practice. If confounders are not observed or not modeled (or even not known), then not only might the magnitude of the estimated treatment effect be biased but it might even have the wrong sign \cite{Hemkens.2018}.

Importantly, assumptions such as those above are required for consistently estimating treatment effects from data, regardless of which method is used. A natural challenge is further that {assessing the plausibility of the assumptions} is often difficult. Later, we discuss potential strategies to check the credibility of whether the assumptions hold. {Notwithstanding, there are also problem setups with other designs. For example, some problem setups allow for relaxations of the SUTVA assumption (e.g., by allowing for spillover effects) \cite{Laffers.2020,Huber.2021}. There also exist alternatives to assuming unconfoundedness in specific settings such as through the use of instrumental variables \cite{Syrgkanis.2019,Frauen.2023iv}. Finally, there are problem setups that are not static but time-varying, so that a sequence of treatment decisions is made over time \cite{Lim.2018,Liu.2020DSW,Li.2021,Bica.2020crn,Liu.2023sepsis,Melnychuk.2022,Schulam.2017,Vanderschueren.2023,Seedat.2022,Hess.2024}. Even other works focus how to effectively combine both observational and experimental data \cite{Hatt.2022combining,Colnet.2020,Kallus.2018removing}.}

\subsection*{Methods}

There are different causal ML methods, which vary based on which causal quantity of interest and which causal model is addressed. {For example, a large body of literature focuses on causal ML for ATE \cite{vanderLaan.2006,vanderLaan.2007, vanderLaan.2011tmle,  Zheng.2011,Diaz.2013,Luedtke.2016}.} For CATE estimation with binary treatments, there are two broader categories of methods (see Box~3). On the one hand, so-called meta-learners \cite{Kunzel.2019} are model-agnostic methods for CATE estimation that can be used for treatment effect estimation in combination with an arbitrary ML model of choice (e.g., a decision tree, a neural network \cite{Curth.2021nonparametric}). On the other hand, model-specific methods make adjustments to existing ML models to address statistical challenges arising in treatment effect estimation and, therefore, to improve performance. Here, prominent examples are the causal tree \cite{Athey.2016} and the causal forest \cite{Wager.2018,Athey.2019}, which adapt the decision tree and random forest, respectively, for treatment effect estimation. Even others adapt representation learning to leverage neural networks for treatment effect estimation \cite{Shalit.2017,Shi.2019}. {In contrast, different methods are needed for continuous treatments such as in settings where the intensity, dosage, or level of exposure to a treatment can be flexibly chosen \cite{Hirano.2004,Bach.2022,Foster.2023, Kennedy.2017, Nie.2021VCNet,Bica.2020scigan, Foster.2023, Hill.2011, Schwab.2020,Schweisthal.2023}.} The reason is that this poses a challenging task as the number of treatment values is infinite and not every value is observed in the data.

{Existing causal ML methods often focus on point estimates. This can be a serious limitation in medical applications \cite{Melnychuk.2023}, where uncertainty estimates such as standard errors or confidence intervals are crucial for reliable decision-making \cite{Banerji.2023}. However, there is also some progress. For example, for CATE estimation, the causal forest \cite{Wager.2018,Athey.2019} is a method that offers rigorous uncertainty estimates. In addition, several other strategies have been developed recently, such as Bayesian methods \cite{Alaa.2017} and conformal prediction \cite{Alaa.2023}, but still more research is needed.}

\noindent\fbox{%
\parbox{\textwidth}{%
\textbf{Box 3. Meta-learners in causal ML}

\begin{flushleft}

Meta-learners \cite{Kunzel.2019} are model-agnostic methods for CATE estimation that can be used for treatment effect estimation in combination with an arbitrary ML model of choice (e.g., a decision tree, a neural network \cite{Curth.2021nonparametric}). There are different ways in which such meta-learners can leverage the data in a supervised learning setting. 

\vspace{0.3cm}

\emph{Plug-in learners:} One way is to train a single ML model that predicts the patient outcome but where the treatment is added as a separate variable to the covariates (called S-learner \cite{Kunzel.2019}). Another way is to train two separate ML models for each treatment (called T-learner \cite{Kunzel.2019}). Here, one ML model is trained for predicting patient outcomes in the treatment group and one ML model for the control group. After having computed the ML model(s), one simply uses the estimated treated and control outcome to ``plug them into'' the formula for computing the treatment effect. 

\vspace{0.3cm}

\emph{Two-step learners:} An alternative approach is to target the CATE, which can lead to faster convergence \cite{Kennedy.2023DR}. However, Because the difference between factual and counterfactual outcomes is never observed in data, so-called pseudo-outcomes are used as surrogates, which have the same expected value as the CATE. {Prominent examples are the so-called DR-learner \cite{Kennedy.2023DR} and the so-called R-learner \cite{Nie.2021quasioracle}, which come with certain robustness guarantees \cite{Chernozhukov.2018, Nie.2021quasioracle, Foster.2023}. } 

\vspace{0.3cm}

{The above meta-learners have different advantages and disadvantages. Unfortunately, there are no clear rules for choosing meta-learners but only high-level recommendations \cite{Curth.2021nonparametric, Morzywolek.2023}.}

\end{flushleft}

}}

\subsection*{Evaluation}

Arguably, the best way to evaluate models is to access randomized data. While this does not allow to assess treatment effects of individual patients, it still helps during model selection, so that models are favored with the best performance in terms of average or heterogeneous treatment effects. In contrast, benchmarking for the purpose of model selection is challenging, as both counterfactuals and ground-truth values of treatment effects are unknown {\cite{Curth.2021benchmarking,Boyer.2023,Keogh.2023}}. As a remedy, two strategies are common. (1)~A simple heuristic is to compare methods from causal ML based only on the performance in predicting factual outcomes, whereby the performance in predicting counterfactual outcomes is ignored. This heuristic may give some insights into whether the underlying disease mechanisms in the data are captured. Yet, it has a major limitation in that the key causal quantity of interest -- i.e., the treatment effect -- is not evaluated. (2)~Another heuristic is to use pseudo-outcomes \cite{Curth.2023modelselection}. Here, a pseudo-outcome is first estimated using a secondary, independent model to approximate the unknown counterfactual outcome, and, then, the pseudo-outcome is used to benchmark the estimated CATE. However, such a heuristic depends on the performance of the secondary model for pseudo-outcomes and tends to favor certain methods \cite{Curth.2023modelselection}. {Still, the two strategies are heuristics and thus have inherent limitations.}

\section*{Technical recommendations}

\subsection*{Checking the plausibility of assumptions}

{Assessing the plausibility of the underlying assumptions} is crucial for the validity of treatment effect estimates, yet also challenging. For the consistency assumption, one should assert that the treatment of one patient does not affect the outcome of another based on domain knowledge. For the positivity assumption, one typically plots the propensity scores to check whether the propensity scores are not too small or too large, since, otherwise, there may not be enough support in the data for reliable inferences \cite{Petersen.2012}. Another strategy is to rely upon methods for uncertainty quantification as some treatments may be given rarely to certain patient cohorts, implying that there may be limited support in the data for making inferences in these patient cohorts and, therefore, a large uncertainty \cite{Jesson.2020}. If the positivity assumption is violated, a strategy is to exclude certain subgroups from the analysis as no reliable inferences for them can be made \cite{Petersen.2012,Rudolph.2022}.

Validating the unconfoundedness assumption is especially challenging for RWD. The best way to avoid violations of the unconfoundedness assumption is to consult domain knowledge to ensure that all relevant factors behind treatment assignment are captured in RWD \cite{Cinelli.2022}. An alternative is to adopt an instrumental variable approach \cite{Syrgkanis.2019,Frauen.2023iv} but valid instruments are often rare in medical applications and, again, the validity of instruments cannot be tested. If unobserved confounders cannot be ruled out, conducting a causal sensitivity analysis can be helpful to assess how robust the results are to potential unobserved confounding. Causal sensitivity analysis dates back to a study from 1959 showing that unobserved confounders cannot explain away the causal effect of smoking on cancer \cite{Cornfield.1959}. Causal sensitivity analysis computes bounds on the causal effect of interest under some restriction on the amount of confounding, thus implying that a treatment effect cannot be explained away. Then, restrictions on the amount of confounding are based on domain expertise, typically by making comparisons to known, important causes that act as baselines (e.g., risk factors such as age). Recently, a series of causal ML methods have been proposed that provide sharp bounds \cite{Frauen.2023sharpbounds,Kallus.2019,Jin.2023,Dorn.2022, Oprescu.2023}. However, causal sensitivity analysis still requires that there is sufficient knowledge of human pathophysiology and pharmacology about important disease causes, which may not always be the case in observational studies \cite{Hemkens.2018}.

In addition, there are so-called refutation methods to validate the robustness of the treatment effect estimates against explicit violations of the different assumptions \cite{Sharma.2021}. Common refutation methods are, for example, adding a random variable to check if the treatment effect estimates remain consistent as such variable should not affect the estimates, or replacing the actual treatment variable with a random variable to check if the estimated treatment effect goes to zero. Further, one could perform simulations where the outcome is replaced through semi-synthetic data to check if the treatment effect is correctly estimated under the new data-generating mechanism for the simulated outcomes. Altogether, the choice of which refutation method to use for validating the used methods highly depends on the specific problem setting and should be carefully chosen and implemented. Still, even when the refutation methods yield a positive result, this is no guarantee that the assumptions are satisfied.

Notwithstanding, robustness checks that are best practice in ML are still essential (e.g., to mitigate the risk of bias \cite{Vokinger.2021}), especially as the results in treatment effect estimation may heavily depend on both the data and the model choice.

\subsection*{Reporting}

Findings should be interpreted with great care. In particular, the assumptions, the rationale for the chosen causal ML method, and the robustness checks should be clearly stated. If possible, the estimated treatment effects from RWD should be compared against those from RCTs. This can help in validating the reliability of the causal ML methods but may also reveal differences between clinical trials and routine care (e.g., due to different patient cohorts or different levels of adherence). The reliability of the estimated treatment effects also depends on the quality and representativeness of the underlying data. Furthermore, analyses through causal ML involve multiple hypotheses testing and, therefore, are at risk of false positives. Similarly, due to the retrospective nature of such analyses, another risk is selective reporting of positive results. To mitigate such risks, preregistered protocols for analysis are highly recommended {\cite{Hernan.2016,Xu.2022}}. Finally, when causal ML is used together with RWD, the limitations of making causal conclusions should be openly acknowledged, and, if possible, RCTs should be considered for validation.

\section*{Clinical translation}

{By estimating treatment effects from medical data, causal ML offers significant potential to personalize treatment strategies and improve patient health. Still, there is a long way to go. A key focus for future research must be on bridging the gap between ML research and direct benefits for patients in clinical practice.} 

\subsection*{Use cases}

{Causal ML can help in generating new clinical evidence. For RCTs, causal ML may determine specific patient cohorts within the population that might respond positively (or negatively) to a particular treatment. For example, the treatment effect of antidepressant drugs over a placebo varies substantially and tends to increase with baseline severity of the depression \cite{Fournier.2010}. However, RCTs typically compare patient outcomes across two (or more) treatment arms, which would return the average treatment effect at the population level, and the use of causal ML may help to define inclusion criteria for clinical trials or to identify predictive biomarkers (e.g., certain genetic mutations in a tumor).} 

{Furthermore, causal ML may offer flexible, data-driven methods to analyze treatment effect heterogeneity in RWD such as clinical registries and electronic health records. This is relevant as RCTs can be subject to limitations \cite{Booth.2019}, such as high costs or that treatment randomization can be unethical for vulnerable populations (e.g., pregnant women) \cite{Chien.2022}. RWD together with causal ML could allow to estimate heterogeneous treatment effects for vulnerable groups, rare diseases, long-term outcomes, and uncommon side effects that are often not sufficiently captured by traditional RCTs. For example, as randomizing hospitalizations is typically not possible, one study used causal ML to estimate the effect of hospitalizations on suicide risk from RWD \cite{Ross.2023}. Likewise, patient populations in RCTs are often not representative of the broader population \cite{Cole.2010}, but which one can account for through causal ML \cite{Hatt.2022opl} in order to better understand the post-approval efficacy of treatments. However, while the potential of RWD has been widely recognized \cite{Sherman.2016, Booth.2019}, many methodological questions are still unanswered, and causal ML may thus help in translating data into clinical evidence.}

{Eventually, the choice of the specific estimand depends on the setting where causal ML is used. For regulatory bodies, it may be relevant to assess the overall net benefit for patients at large, for example, when comparing a new drug against the standard of care. This would require the estimation of the average treatment effect. To ensure patient safety, regulatory bodies could also assess how the treatment effect varies across different subpopulations, which would involve the conditional average treatment effect. Likewise, the conditional average treatment effect may help to identify subpopulations that are particularly responsive to a treatment (e.g., for hypothesis generation) or that would benefit from newly developed drugs, thereby contributing to an accelerated drug development. When causal ML is integrated into clinical decision support systems in routine care, clinical professionals may want to make personalized predictions of how a patient's health state changes under different treatment options. This would require methods for CATE estimation or even for predicting potential patient outcomes.}

\subsection*{Challenges and future directions}


{Several challenges in the clinical translation of causal ML are at the technical level. First, both estimating heterogeneous treatment effects as well as predicting patient outcomes are naturally difficult. In practice, this often requires both strong predictors of treatment effects and large sample sizes. While the former depends on the human pathophysiology and pharmacology in the specific disease setup, the latter may improve over time with an increasing prevalence of electronic health records. Another challenge is that uncertainty quantification for many causal ML methods is lacking. However, uncertainty quantification is crucial for reliable decision-making and thus for building clinical evidence \cite{Banerji.2023}. For example, point estimates might indicate substantial effect heterogeneity, especially in settings with limited data, while, in fact, there is little heterogeneity but simply large (aleatoric) uncertainty as the outcomes are difficult to predict. Hence, causal ML methods that only provide point estimates without conveying the appropriate uncertainty in the predictions may lead to potentially misleading or inappropriate conclusions. Finally, many causal ML methods are only implemented in specialized software libraries. Hence, comprehensive software tools are needed that improve reliability and ease of use and account for needs in medicine (e.g., rigorous uncertainty quantification).}

The development of standardized protocols, ethical guidelines, and regulatory frameworks for causal ML applications will be essential in ensuring safe and effective treatment decisions. For example, tailored checklists based on consensus statements will need to be developed. While there are checklists for traditional, predictive ML \cite{Norgeot.2020} and for generating real-world evidence {\cite{vonElm.2007,Xu.2022}}, future research is needed that adapts such checklists to account for the needs of causal ML in medicine. Likewise, customized review processes will need to be developed, which define how evidence generated through the causal ML method must undergo regulatory review for approval.

{So far, research in causal ML has primarily evaluated the performance of different methods through simulations (e.g., \cite{Lim.2018,Bica.2020crn,Melnychuk.2022,Li.2021,Vanderschueren.2023,Hess.2024}). However, simulations involve (semi-)synthetic datasets that do not fully capture the nuances of real-world disease dynamics. Hence, demonstrating the clinical insights generated through a cautious use of innovative causal ML methods can provide an important first step.} This will help in understanding the strengths and limitations of causal ML in a medical context, especially in comparison to established clinical trials. For this, settings may appear especially suited where clear guidelines are missing, so that clinical decision support through causal ML can provide additional input by augmenting the decision-making of clinical professionals. {Eventually, tools based on causal ML may be integrated into routine care through clinical decision support systems. Such systems may directly predict individual patient outcomes for different treatment options and thereby support the decision-making of clinical professionals.}

\section*{Conclusion}

{Causal ML offers significant potential to draw novel conclusions about the efficacy and safety of treatments. In particular, causal ML offers abundant opportunities to personalize treatment strategies and thus to improve patient health. However, several challenges arise in practice. One challenge is to ensure the reliability and robustness of causal ML methods. Another challenge is to overcome barriers in the clinical translation. Here, proof-of-concept studies and a cautious use in practice can be an important first step.} 

\section*{Acknowledgments}

SF acknowledges funding via the Swiss National Science Foundation (SNSF), Grant 186932.

\vspace{0.4cm}
\section*{Author contributions} 

All authors contributed to conceptualization, manuscript writing, and approved the manuscript.

\vspace{0.4cm}
\section*{Competing interests}
The authors declare no competing interests.


\newpage
\bibliography{literature}

\vspace{1cm}
\noindent
This version of the article has been accepted for publication, after peer
review but is not the Version of Record and does not reflect post-acceptance improvements, or any corrections. The Version of Record is available online at: \url{https://doi.org/10.1038/s41591-024-02902-1}. Use of this Accepted Version is subject to the publisher’s Accepted Manuscript terms of use https://www.springernature.com/gp/open-research/policies/acceptedmanuscript-terms

\end{document}